\documentclass[final,5p,times,twocolumn]{elsarticle}

\usepackage{dcolumn}
\usepackage[mathscr]{euscript}

\usepackage{xcolor}
\usepackage{listings}
\usepackage{amsmath}
\usepackage{hyperref}

\definecolor{dkgreen}{rgb}{0,0.6,0}
\definecolor{gray}{rgb}{0.5,0.5,0.5}
\definecolor{mauve}{rgb}{0.58,0,0.82}

\let\counterwithin\relax
\usepackage{chngcntr}
\AtBeginDocument{\counterwithin{lstlisting}{section}}

\usepackage{todonotes}

\newcolumntype{d}[1]{D{.}{.}{#1}}

\begin{document}
\begin{frontmatter}

\title{SOSA: A Lightweight Ontology for Sensors, Observations, Samples, and Actuators} %IOT

\author[a]{Krzysztof Janowicz\corref{cor1}}
\ead{janowicz@ucsb.edu}
\author[b]{Armin Haller}
\ead{armin.haller@anu.edu.au}
\author[c]{Simon J D Cox}
% Yes, I know the J D looks like an affectation, but the rest of the name is far from unique in Computer Science or Geology
\ead{simon.cox@csiro.au}
\author[d]{Danh Le Phuoc}
\ead{danh.lephuoc@tu-berlin.de}
\author[e]{Maxime Lefran\c{c}ois}
\ead{maxime.Lefrancois@emse.fr}

\address[a]{Department of Geography, University of California, Santa Barbara, USA}
\address[b]{Research School of Computer Science, Australian National University, Canberra, Australia}
\address[c]{Land and Water, CSIRO, Melbourne, Australia}
\address[d]{Department of Telecommunication Systems, Technische Universit{\"a}t, Berlin, Germany}
\address[e]{Connected-Intelligence team, Ecole des Mines de Saint-Etienne, France}

%\runningtitle{SOSA: A Lightweight Ontology for Sensors, Observations, Samples, and Actuators}

%\author[A]{\inits{A.}\fnms{Krzysztof} \snm{Janowicz}\ead[label=e1]{janowicz@ucsb.edu}%
%\thanks{Corresponding author. \printead{e1}.}},
%\author[B]{\inits{K.}\fnms{Armin} \snm{Haller}\ead[label=e2]{armin.haller@anu.edu.au}},
%\author[C]{\inits{S.}\fnms{Simon} \snm{Cox}\ead[label=e3]{simon.cox@csiro.au}},
%\author[D]{\inits{D.}\fnms{Danh} \snm{Le Phuoc}\ead[label=e4]{danh.lephuoc@tu-berlin.de}}, and
%\author[E]{\inits{M.}\fnms{Maxime} \snm{Lefran\c{c}ois}\ead[label=e6]{maxime.Lefrancois@emse.fr}}
%\runningauthor{K. Janowicz et al.}

%\address[A]{Geography Department, \institution{University of California},
%Santa Barbara, CA, \cny{USA}\printead[presep={\\}]{e1}}
%\address[B]{Research School of Computer Science, \institution{Australian National University}, Canberra, \cny{Australia}\printead[presep={\\}]{e2}}
%\address[C]{Land and Water, \institution{CSIRO},
%Melbourne, \cny{Australia}\printead[presep={\\}]{e3}}
%\address[D]{Department of Telecommunication Systems, \institution{Technische Universit{\"a}t},
%Berlin, \cny{Germany}\printead[presep={\\}]{e4}}
%\address[E]{Connected-Intelligence team, \institution{Ecole des Mines de Saint-Etienne}
%Saint-\'{E}tienne, \cny{France}\printead[presep={\\}]{e6}}

\date{April 9, 2018}

\begin{abstract}
 The Sensor, Observation, Sample, and Actuator (SOSA) ontology provides a formal but lightweight general-purpose specification for modeling the interaction between the entities involved in the acts of observation, actuation, and sampling. SOSA is the result of rethinking the W3C-XG Semantic Sensor Network (SSN) ontology based on changes in scope and target audience, technical developments, and lessons learned over the past years. SOSA also acts as a replacement of SSN's Stimulus Sensor Observation (SSO) core. It has been developed by the first joint working group of the Open Geospatial Consortium (OGC) and the World Wide Web Consortium (W3C) on \emph{Spatial Data on the Web}. In this work, we motivate the need for SOSA, provide an overview of the main classes and properties, and briefly discuss its integration with the new release of the SSN ontology as well as various other alignments to specifications such as OGC's Observations and Measurements (O\&M), Dolce-Ultralite (DUL), and other prominent ontologies. We will also touch upon common modeling problems and application areas related to publishing and searching observation, sampling, and actuation data on the Web. The SOSA ontology and standard can be accessed at \url{https://www.w3.org/TR/vocab-ssn/}.
\end{abstract}

\end{frontmatter}
\sloppy

\section{Introduction and Motivation}
% Following SOSA's philosophy, this is written for a broad audience, not just semantic webbers.

%\todoah{R3: The procedure class appears to be key to ensuring results of actions are interpreted correctly, yet it is unclear what the intended scope of the class is from this article. Is it essentially the required deployment instructions ("sensor must be located X distance from the ground"), what was actually performed ("sensor was located next to a breezy window"), how the sensor/actuator performed the action ("by applying a formula to convert an input signal into a result value"), or all of these and more? Note - the ontology itself is slighly about this.}
%RESPONSE: We have included a paragraph in the Procedures section that details the difference between Procedures for Observations and Sampling and Procedures for Actuations.

%\todoall{R2: There is also a slight confusion, at least in the first few sections of the paper, about the paper focus. At the beginning I get the impression that although the paper focus is on SOSA, the reader will get an overview of SSN, and how it builds on SOSA, its alignments to DUL and other ontologies. Differences in expressivity between the modules are discussed, etc. However, later in the paper it turns out that only SOSA is presented, and just a few sentences about its extension and alignments are included in the end. While I think this is the exact right focus of the paper, this can be made more clear at the beginning of the paper - SOSA is the focus, anything else is out of scope more or less.}

In their broadest definition sensors detect and react to changes in the environment that directly or indirectly reveal the value of a property.  The process of determining this, not necessarily numeric, value is called an observation. Observation procedures provide a sequence of instructions to ensure that the observations are \emph{reproducible} and \emph{representative}, whereby an individual assessment characterizes a feature (i.e., entity) of interest. Typically, observations are not carried out on the entire feature but on samples of it, or on an immediately sensed spatiotemporal region. The process of sampling may itself be specified by a procedure that determines how to obtain samples. Some observation procedures can contain sampling procedures as their parts. Actions triggered by observations are called actuations and the entities that perform them are actuators. Finally, actuators, sensors, and samplers are typically mounted on a platform. These platforms serve a wide range of needs, including carrying systems along a defined trajectory, protecting them from external influences that may distort the results, or spatially positioning multiple systems following a particular layout. 

In the context of smart homes, for instance, a temperature sensor can be mounted to a wall and take repeated observations at some time interval. Each of these observations returns the temperature of a sample, namely the surrounding body of air. In cases where the sensor is placed correctly, the temperature is said to be characteristic (representative) for the entire feature of interest, e.g., a bedroom. For example, a decrease in room temperature may trigger an actuator to close the windows. While each individual observation results in a new value and is taken from a new sample, all observations are based on the same procedure, observe the same property, and reveal the same characteristic of the same feature of interest. Ignoring some aspect of the observation procedure - e.g., by placing the sensor next to the window so that the sampled body of air is no longer a suitable proxy for the entire room - may cause the observation results to become unrepresentative of the room's temperature, and may lead to the actuator closing the window unexpectedly. Note that if one would place all sensors in such way, the resulting observations may not be representative for the room though they still may be reproducible. Finally, if some sensors would be placed near windows while others would not, it would no longer be possible to establish a relationship between the behavior of individual actuators. Finally, procedures are specific for certain types of observations. Hence, one can follow a specific procedure and thereby arrive at reproducible results that is not representative or suitable for the task at hand.

With a rapid increase in data from sensors being published on the Web, there is an increasing interest in the re-use and combination of that data. However, raw observation results do not provide the context required to interpret them properly and to make sense of these data.   Searching, reusing, integrating, and interpreting data requires more information about the studied feature of interest, such as a room, the observed property, such as temperature, the utilized sampling strategy, such as the specific locations and times at which the temperature was measured, and a variety of other information. With the rise of smart cities and smart homes as well as the Web of Things more generally, actuators and the data that is produced by their in-built sensors also become first-class citizens of the Web. Given their close relation to sensors, observations, procedures, and features of interest, outlined above it is desirable to provide a common framework and vocabulary that also includes actuators and actuation. Finally, with today's diversity of data and data providers, notions that restrict the view of sensors to being technical devices need to be broadened. One example would be social sensing techniques  such as  semantic signatures \cite{odoe} to study humans and the data traces they actively and passively emit from within a sensor-observation framework. Simulations and forecasts are other examples showcasing why `sensors' that produce estimates of properties in the world are not necessarily physical entities.

The Sensor Web Enablement standards such as the Observations and Measurements (O\&M) \cite{OaM} model and the Sensor Model Language  (SensorML) \cite{sensorml} specified by the Open Geospatial Consortium (OGC) provide means to annotate sensors and their observations. However, these standards are not integrated and aligned with Semantic Web technologies, Linked Data, and other parts of the World Wide Web Consortium's (W3C) technology stack that aims at creating and maintaining a global and densely interconnected graph of data. The W3C Semantic Sensor Network Incubator Group (SSN-XG) tried to address this issue by first surveying the landscape of semantically-enabled sensor specifications~\cite{compton2009survey} and then developing the Semantic Sensor Network (SSN) ontology~\cite{ssn} as a human and machine readable specification that covers networks of sensors and their deployment on top of sensors and observations. To provide an axiomatization beyond mere surface semantics, SSN made use of the foundational Dolce UltraLight (DUL) ontology, e.g., to state that platforms are physical objects. At the same time, SSN also provided the Sensor-Stimulus-Observation (SSO) \cite{janowicz2010stimulus} ontology design pattern \cite{gangemi2009ontology} as a simple core vocabulary targeted towards lightweight applications and reuse-by-extension.

\begin{figure}[ht]
\centering
  \includegraphics[width=\linewidth]{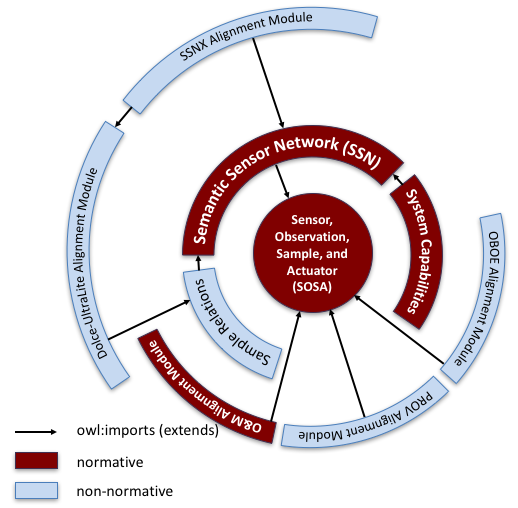}
\caption{SOSA and its vertical and horizontal modules with the arcs indicating the direction of the import statement. Horizontal modularization is shown by arcuate modules at the same radius while vertical modularization is shown by modules at a larger radius.}
\label{fig:modules}
\end{figure}

%\todoall{R1: It is clear why the discussion about the modular structure as well as the relation with the other existing ontologies is needed, but I found it too detailed for the general purpose of the introduction.}
The broad success of the initial SSN led to a follow-up standardization process by the first joint working group of the OGC and the W3C. One of the tasks of this \emph{Spatial Data on the Web} working group was to rework the SSN ontology based on the lessons learned over the past years and more specifically to address changes in scope and audience, shortcomings of the initial SSN, as well as technical developments and trends in relevant communities. The resulting ontology, published as a W3C Recommendation and OGC Standard~\cite{ssn:17}, is not only an update but has been re-envisioned completely from the beginning. Most notably, and as depicted in Fig.~\ref{fig:modules} the revised ontology is based on a novel modular design which introduces a horizontal and vertical segmentation. Vertical modules add additional depth to the axiomatization by directly importing lower modules and defining new axioms, while horizontal modules broaden the ontology's scope, e.g., by introducing classes and relations to specify system capabilities or sample relationships, but do not otherwise enrich the semantics of existing terms. The modularization addresses an often voiced concern about the initial SSN release, that the DUL alignment introduced too strong ontological commitments, and the full ontology was too heavyweight for smart devices in the context of the Web of Things, and was running against the trend towards lightweight vocabularies preferred by the Linked Data and Schema.org communities. The proposed modularization allowed us to keep the DUL alignment for those who want to use it, and introduce additional alignments to Prov-O \cite{lebo2013prov}, O\&M \cite{OaM}, and OBOE \cite{madin2007ontology}, while keeping the overall target audience broad, ranging from web developers and scientists that want to publish their data on the Web, to Web of Things industry players.

%\todoall{R3: Fig 1 and the associated text mention vertical and horizontal alignments of SOSA with other ontologies; however, it is unclear which alignments are vertical and which are horizontal from the layout of the figure.  Section 3 implies that all of these alignments are vertical.  Please clarify.}
%RESPONSE: We included a description of horizontal and vertical modularization in the caption of the figure.

The resulting collection of modules, including SSN, all build upon a common core: the \textbf{S}ensor,  \textbf{O}bservation,  \textbf{S}ample, and  \textbf{A}ctuator ontology (SOSA). SOSA does not merely replace the former SSO ontology design pattern but provides a flexible yet coherent framework for representing the entities, relations, and activities involved in sensing, sampling, and actuation. It is intended to be used as a lightweight, easy to use, and highly extendable vocabulary that appeals to a broad audience beyond the Semantic Web community but can be combined with other ontologies, such as SSN to provide a more rigorous axiomatization where needed. At the same time, SOSA acts as minimal interoperability fall-back level, i.e., it defines those common classes and properties for which data can be safely exchanged across all uses of SSN, its modules, and SOSA.

In the following, we will focus on providing an overview of the main classes and properties of SOSA. We will also briefly discuss their integration with the new SSN ontology. We will motivate some of the core design decisions and provide modeling examples that will arise in practice. For the sake of readability, we will focus on an examples-driven description of these classes. The formal and normative SOSA ontology and standard can be accessed at \url{https://www.w3.org/TR/vocab-ssn/}. Finally, we will discuss selected modeling problems and how to approach them and will give examples for the usage of SOSA classes and relationships in different application areas.

\section{SOSA in a Nutshell}

%\todoall{R3: I appreciate that the working group conducted signification work in redesigning the initial SSN-XG ontology, including accommodating feedback from users.  Section 1 and 2 both mention that the initial SSN-XG was based around the SSO design pattern, and the revised SOSA (and SSN) ontology adopt an event-centric perspective, yet the reasons for this change are not discussed in the article - any reader familiar with SSN-XG but not SOSA is left wondering why this change was made.  Please add some context regarding the main reasons for it, and the implications for anyone who has been using the SSN-XG ontology in terms of how to ensure their data are interpretable with SOSA.}
%RESPONSE: We have provided references and evidence for why the new event-centric modelling is aligned with the We community's expectations, i.e. in particular the way Schema.org models actions, events in PROV-O and events in all types of business process-based Web service interfaces.

%\todoall{R3: Is the core structure ontology design pattern unique to SOSA or based on pre-existing pattern(s)?}

%\todoah{R3: Paragraph 1 mentions alignment with Prov-0 - this is the first time Prov-O is mentioned, and it is unclear at this point why this alignment is desirable - please add some discussion explaining why this alignment was made.}
%RESPONSE: We have removed the reference in paragraph 1, but included more detail how the activity-centric view of SOSA is more aligned with existing standards, among them PROV-O in the subsequent paragraph where we describe Figure 2.

Here, we will highlight the most important classes and relationships that make up the SOSA ontology. In contrast to the original SSN, SOSA takes an event-centric perspective and revolves around observations, sampling, actuations, and procedures. The last is a set of instructions specifying how to carry out one of the three aforementioned acts. This event-centric modelling is aligned with community expectations, in particular the Schema.org community that only cares about the digital representation of an event (i.e. Action class in the Schema.org model)~\cite{schemaorg} and not the real-world process that underlies such an event. Modelling events as first-class citizens is also aligned with the PROV-O model~\cite{lebo2013prov} and standardized business process models that are used for Web service interfaces~\cite{vanderaalst}. 

\begin{figure}[ht]
\centering
  \includegraphics[width=\linewidth]{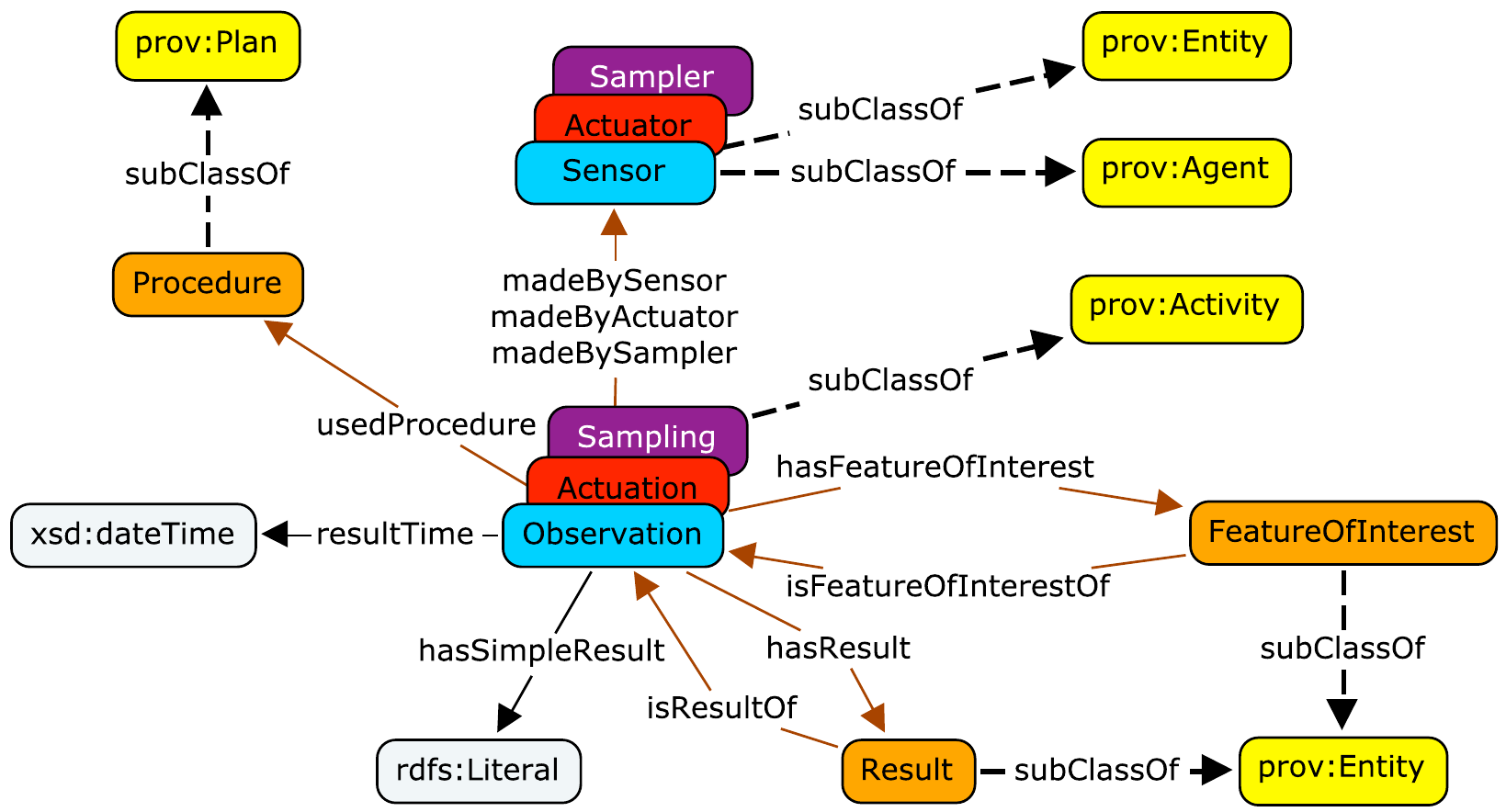}
\caption{The core structure of SOSA, showing relationship to Prov-O}
\label{fig:core_structure}
\end{figure}

Fig.~\ref{fig:core_structure} depicts the SOSA ontology design pattern, called \emph{core structure}, underlying all of the three modeling perspectives. The \textit{activities} of observing, sampling, and actuating, each target some \textit{feature of interest} by either changing its state or revealing its properties, each follows some \textit{procedure}, and each is carried out by some object or \textit{agent}. The core structure aligns well with other activity-centric standard ontologies such as the OGC Observations and Measurements~\cite{om:11} (i.e. \texttt{Observation} $\equiv$ \texttt{so19156-om:OM\_Observation}) and the PROV Ontology~\cite{lebo2013prov} (i.e. \texttt{Observation} $\sqsubseteq$ \texttt{prov:Activity}). Fig.~\ref{fig:core_structure} shows how some of the core classes in SOSA relate to an activity-centric model such as PROV-O. Note that SOSA does not model the ultimate agent that triggered an act, e.g., a person taking a reading off a sensor. We will show an example how this agent relationship can be modelled by using PROV-O in combination with SOSA \mbox{in Sec.~\ref{sosaprovo}.}

% SC: don't like the use of prov: prefixed labels in the figure. This elides too many intermediate steps, and doesn't actually represent any real graph. 
% The primary goal of this figure is to show the core pattern, and alignment with prov-o is a secondary consideration at this point of the narrative
% I think it would be better to drop the prov: prefixes, and maybe just change the font of the non-sosa terms to italics. 
% then point out the alignment with PROV in the caption or in the text. 

SOSA aims to strike a balance between the expressivity of the underlying description logic, the ease of use of language features, e.g., as measured by understanding their implications, cf. \cite{rector2004owl,horridge2011cognitive},  and the expectations of the target audience (including web developers and domain scientists), while accommodating a broad range of domains and application areas. Given SOSA's axiomatization, the resulting DL language fragment is $\mathcal{ALI(D)}$ which is efficiently supported by modern triple stores. In comparison the new SSN module's axiomatization results in $\mathcal{ALRIN(D)}$, while the old SSN was specified in $\mathcal{SRIQ}$. 

To give a concrete example,  SOSA does not declare classes to be disjoint despite this being among the most powerful language elements in terms of reasoning. This is for good reasons as it enables classes such as sensors and samples to be features of interest themselves. Consequently, one can make observations \textit{using} sensors as well as \textit{about} sensors. Similarly, features of interest can be regarded as the results of a sampling activity. Another notable design decision is the usage of Schema.org \emph{domainIncludes} and \emph{rangeIncludes} annotation properties~\cite{schemaorg} to provide an informal semantics, compared to the inferential semantics of their RDFS counterparts. 
The SSN vertical module continues to utilize \emph{guarded} domain and range restrictions as part of its richer axiomatization.

%\begin{figure}%
%\centering
%\subfigure[][]{%
%\label{fig:sosa-cpo}%
%\includegraphics[height=2in]{./figures/observation.png}}%
%\hspace{8pt}%
%\subfigure[][]{%
%\label{fig:sosa-cpa}%
%\includegraphics[height=2in]{./figures/actuation.png}} \\
%\subfigure[][]{%
%\label{fig:sosa-cps}%
%\includegraphics[height=2in]{./figures/sampling.png}}%
%\hspace{8pt}%
%\subfigure[][]{%
%\label{fig:sosa-cpp}%
%\includegraphics[height=2in]{./figures/prov-o.png}}%
%\caption[Common core structure]{The core structure of SOSA's main components as they relate to Prov-O:
%\subref{fig:sosa-cpo} Observation;
%\subref{fig:sosa-cpa} Actuation;
%\subref{fig:sosa-cps} Sampling; and,
%\subref{fig:sosa-cpp} Prov-O.}%
%\label{fig:sosa-cp}%
%\end{figure}

\subsection{Procedures}
%\todoah{Paragraph 4 - what are procedures "crucial for semantic interoperability" of? Is it observations, samples, and actuations (isn't that the role of the observation, sample, actuation classes), or is it the interpretation of these (and related) data?}
%RESPONSE: We have added a sentence outlining how procedures help interoperability of observation or sampling by recording/identifying the activities that were performed.

 A \texttt{Procedure} is a workflow, protocol, plan, algorithm, or computational method that specifies how to carry out an observation, collect a sample, or make a change to the state of the world via an actuator. A procedure is re-usable in the sense that it is executed whenever one performs observations, actuations or samplings of a certain type, thereby making the results reproducible. Hence, procedures, are also crucial for fostering semantic interoperability~\cite{odoe} as \texttt{Procedures} linked to observation and sampling activities are typically a record of how these activities are/were performed. Procedures linked to actuation activities, however, can either be a record of how the actuation has been performed or a description of how to interact with an actuator (i.e., the recipe for performing actuations). The definition of the inputs and outputs of a procedure that define how to interact with an \texttt{Actuator} (i.e. its interface) are beyond the scope of SOSA and are relegated to SSN in combination with other standards/models such as the Linked Data Platform protocol~\cite{ldp} or the currently under development W3C Thing Description~\cite{thingdesc}.
 
 To give a concrete example of the modelling of \texttt{Procedures}, the measured wind speed differs depending on the height of a sensor above the surface, e.g., due to friction. Hence, procedures for measuring wind speed define a standard height for anemometers above ground, typically 10m for meteorological measures and 2m in Agrometeorology. This definition of height, sensor placement, avoidance of obstructions, etc., as expressed in the \lstinline$<AgrometeorologyWindSpeed>$ procedure in the example in Listing~\ref{list_reusable_procedures} make the observation results reproducible and representative. Multiple successive observations will yield comparable results and the measured values will be representative for the feature of interest, e.g., wind speed at a certain stretch of California's Central Coast. While the number of (observation) acts increases constantly, any change in procedure (e.g., an increase in the number of used procedures) would be unexpected and may point to a novel technique, theory, or instrument.
 
 \begin{lstlisting}[caption={Reusable procedures},label=list_reusable_procedures]
 <AgrometeorologyWindSpeed> a sosa:Procedure ;
   rdfs:comment "Instructions for measuring wind speed 2m above ground surface for applications in Agrometeorology."@en ;
   [...].
   
 <AWSObs1> a sosa:Observation ;
   [...]
   sosa:usedProcedure <AgrometeorologyWindSpeed> .
   
 <AWSObs2> a sosa:Observation ;
   [...]
   sosa:usedProcedure <AgrometeorologyWindSpeed> .
  \end{lstlisting}

\subsection{Sensors and Observations}

 An \texttt{Observation} is the act of carrying out an (observation) \texttt{Procedure} in order to estimate or calculate a value of an \texttt{ObservableProperty} of a \texttt{FeatureOfInterest} or a \texttt{Sample} thereof. An observation involves a \texttt{Sensor} and yields a \texttt{Result}. While SOSA relies on QUDT~\cite{qudt} or other vocabularies to describe observation results and their values, an additional datatype property is provided to handle the simple case that merely requires a typed literal, via the \texttt{hasSimpleResult} property. The \texttt{hasResult} object property allows one to model \texttt{Results} as first class citizens and make statements about them by, for example, stating the unit of measurement for the value. Listing~\ref{list_sample} shows an example of the use of the \texttt{hasSimpleResult} for an \texttt{Observation}, while Listing~\ref{list_actuation} models a \texttt{Result} object with a unit of measurement defined for the object.
 
 Sensors in SOSA can be physical devices, but also simulations, numerical models or people, to give a few examples. Sensors respond to stimuli, e.g., a change in the environment, or input data composed from the results of prior observations, to generate the result. Conceptually, they can be thought of as implementations (of parts) of an observation procedure. One or more sensors (as well as actuators and samplers) can be hosted on a \texttt{Platform}. Such platforms can also define the geometric properties, i.e., placement, of sensors in relation to one another. SOSA does not provide explicit properties to model the geometry, but delegates this to ontologies dealing with space, metric and topological properties, e.g., GeoSPARQL~\cite{perry2012ogc}. SOSA also distinguishes between \texttt{phenomenonTime} and \texttt{resultTime}. The former is the time that the result of an act of observation, actuation, or sampling applies to the feature of interest, while the latter specifies the \emph{instant} of time when the act was completed. Consequently, \texttt{resultTime} is a datatype property, while \texttt{phenomenonTime} - which may be an interval or an instant - is an object property which utilizes the OWL-Time vocabulary\footnote{\url{https://www.w3.org/TR/owl-time/}}.  
 
\begin{figure}[ht]
\centering
  \includegraphics[width=\linewidth]{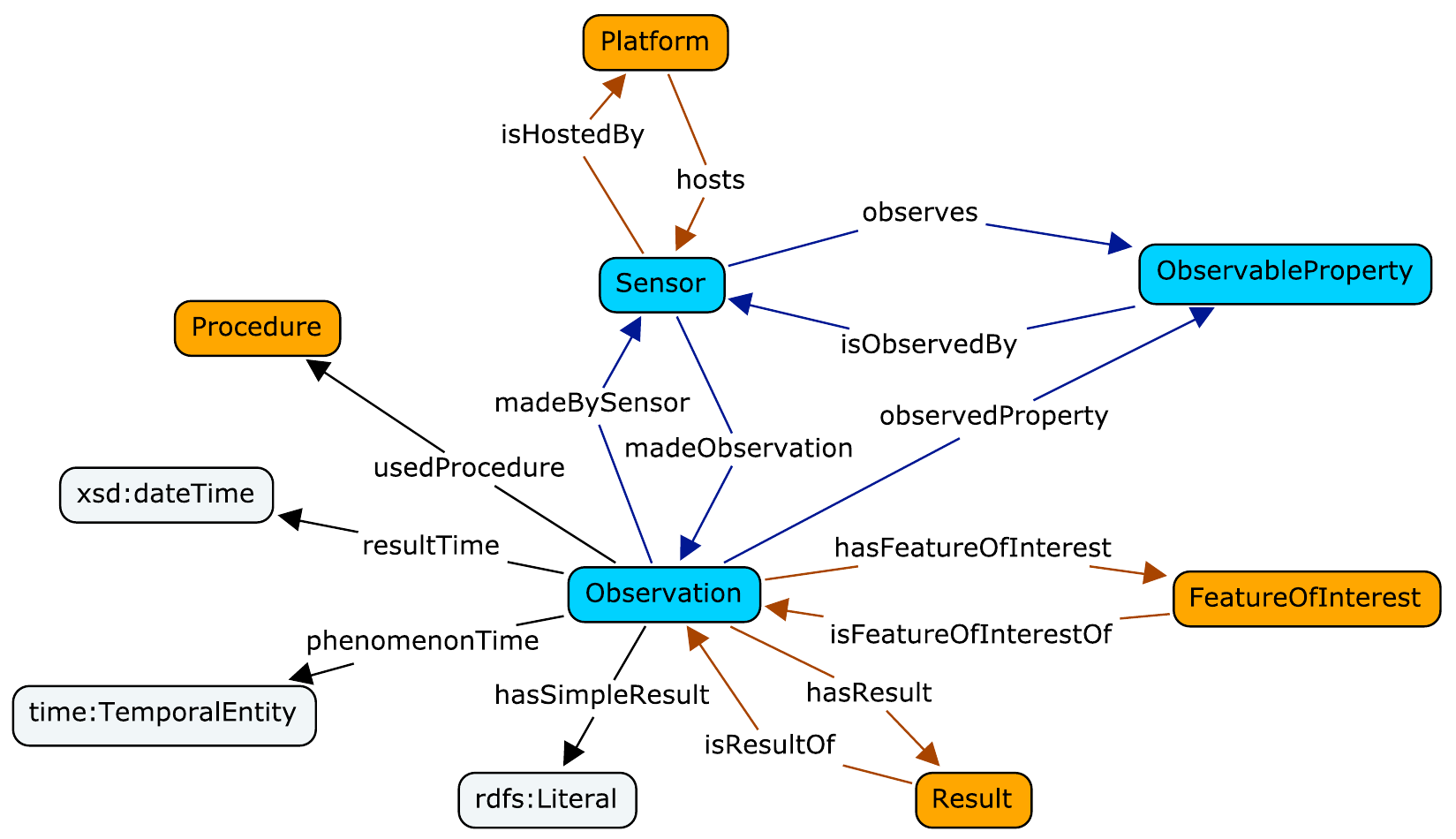}
\caption{Overview of the SOSA Observation perspective}
\label{fig:observation}
\end{figure}

 Finally, observable properties are similar to procedures or units of measure in the sense that they are singletons; one observable property will apply to many acts of observation, concerning different features of interest, at different times, or using different procedures or sensors, e.g., \lstinline$<NitrateConcentration>$ of rainwater from different storms as in Listing~\ref{list_reusable_properties}. Hence, to foster interoperability and reproducibility, we expect observable properties (and procedures) to come from controlled vocabularies based on code lists typically used in the sciences, e.g., SDN:P01::RNCNIT09\footnote{``Concentration of nitrate {NO3- CAS 14797-55-8} per unit volume of rainwater [dissolved plus reactive particulate \textless 0.4/0.45um phase] by ion chromatography"; URI  \url{http://vocab.nerc.ac.uk/collection/P01/current/RNCNIT09/}.}. Note that the SOSA axiomatixation does not make any formal restrictions on how to use observable properties, thus allowing for alternative models. Features of interest are not restricted to objects such as trees but also include events, e.g., a storm. In fact, sensors themselves can be features of interest for other sensors. For instance, an ophthalmologist uses sensors to examine a human's eye. Fig.~\ref{fig:observation} depicts the discussed classes and their relations.

 \begin{lstlisting}[caption={Reusable Observable Properties},label=list_reusable_properties]
 <NitrateConcentration> a sosa:ObservableProperty .
 <NCR_Ion_Chromatography_Procedure> a sosa:Procedure.
 <MetrohmXYZIonChromatographySystems> a sosa:Sensor. 
   
 <NC_S1> a sosa:Observation ; [...]
  sosa:usedProcedure <AgrometeorologyWindSpeed> ;
  sosa:madeBySensor  <MetrohmXYZIonChromatographySystems>;
  sosa:observedProperty <NitrateConcentration>.
   
 <NC_S2> a sosa:Observation ; [...]
   sosa:observedProperty <NitrateConcentration> .
  \end{lstlisting}

One or more sensors, actuators or samplers can be hosted or mounted on a \texttt{Platform}. Such platforms provide, for example, geometric properties essential for performing observations, actuation or sampling, e.g., by placing the microphone at a certain position and away from the speakers, the flash pointing in the same direction as the rear-facing camera, and so on. \texttt{Platforms} can also host other \texttt{Platforms}.

Increasingly, accelerometers, linear actuators, gyroscopes, barometers, magnetometers, microphones, cameras and other sensors and actuators are mounted on modern smartphones, which can be modelled as platforms in SOSA. Listing~\ref{list_platforms} defines an IPhone 7 as the platform that hosts a Bosch Sensortec BMP282, a taptic engine linear actuator and a GPS sensor. The listing also shows how the location of the smartphone measured by the GPS sensor with its latitude and longitude (and expressed through the Geo Vocabulary~\footnote{See \url{https://www.w3.org/2003/01/geo/}}) at a given time can be modelled as an observable property to an observation. Other than to an observation, spatial location properties can also be attached to features of interest, platforms, sensors, actuators or samplers to allow both, the modelling of static and remote or mobile sensing/actuation/sampling.

\begin{lstlisting}[caption={Platform modelling},label=list_platforms]
<iphone7/35-207306-844818-0> a sosa:Platform ;
  rdfs:label "IPhone 7 - IMEI 35-207306-844818-0"@en ;
  sosa:hosts <sensor/35-207306-844818-0/BMP282> ;
  sosa:hosts <actuator/35-207306-844818-0/tapticEngine> ;
  sosa:hosts <sensor/35-207306-844818-0/gps> .
  
<sensor/35-207306-844818-0/gps> a sosa:Sensor ;
  sosa:madeObservation <35-207306-844818-0/location/1> .
  
<35-207306-844818-0/location/1> a sosa:Observation ;
  sosa:observedProperty <location> ;
  sosa:resultTime
    "2017-08-18T00:00:12+00:00"^^xsd:dateTimeStamp ;
  sosa:hasResult [
   a geo:Point ;
   geo:lat "51.5"^^xsd:decimal ;
   geo:long "-0.12"^^xsd:decimal ;
   ] .

<actuator/35-207306-844818-0/tapticEngine> a sosa:Actuator ;
  sosa:actsOnProperty <tactileFeedback> ;
  sosa:usedProcedure <UIImpactFeedbackGeneratorAPI> .
\end{lstlisting}

Listing~\ref{list_sample} shows how to model an observation involving a sample that is representative of the atmospheric pressure of Hurricane Maria at a given time. A \texttt{Sample} in SOSA is a \texttt{FeatureOfInterest} itself but also a \texttt{Result} of the act of \texttt{Sampling} as will be discussed in more detail below.

\begin{lstlisting}[caption={Sample modelling},label=list_sample]
<HurricaneMaria> a sosa:FeatureOfInterest ;
  rdfs:label "Hurricane Maria, 2017 season"@en .
<HurricaneMariaAPSampleAtStation1> a sosa:Sample;
  sosa:isSampleOf <HurricaneMaria>.

<Obs123> a sosa:Observation ;
  [...]
  sosa:featureOfInterest <HurricaneMariaAPSampleAtStation1> ;
  sosa:observedProperty <AtmosphericPressure> ;
  sosa:hasSimpleResult "101000 Pa"^^cdt:pressure ;
  sosa:resultTime "2017-09-19T23:00:00Z"^^xsd:dateTime .
\end{lstlisting}

%\todoah{R2: At the end of section 2.2 a datatype is discussed. It is not clear whether this is something that is currently not supported, but should be in the opinion of the authors, or if it is something that is commonly used and usually is supported by RDF stores and SPARQL engines.}
%RESPONSE: We have clarified that these datatypes are not supported by SPARQL and RDF engines, but we feel that they should.

The observation described above uses a custom datatype~\cite{cdt} (i.e. \texttt{cdt:pressure}) that leverages the Unified Code of Units of Measures, a code system intended to include \emph{all} units of measures being contemporarily used in international science, engineering, and business. Such custom datatypes, although compatible with the RDF specification, are not yet recognized datatype IRIs and therefore not supported by current RDF and SPARQL engines. For Web applications we feel that such datatypes are useful and their support would allow an easy comparison of quantity values.

\subsection{Samples, Samplers, and Sampling}

%\todosc{R2: From section 2.3 I get the impression that a sampling is a kind of observation, i.e. a subclass. Is this the case? Why? Why not?}

%\todosc{R3: The definition of a Sampler is that it's a "device that implements a sampling procedure" - it is unclear if the ontology places any restrictions on the nature of that device - for example, is the sampler the thermal drill or the person operating it? Please clarify.}

\begin{figure}[ht]
\centering
  \includegraphics[width=\linewidth]{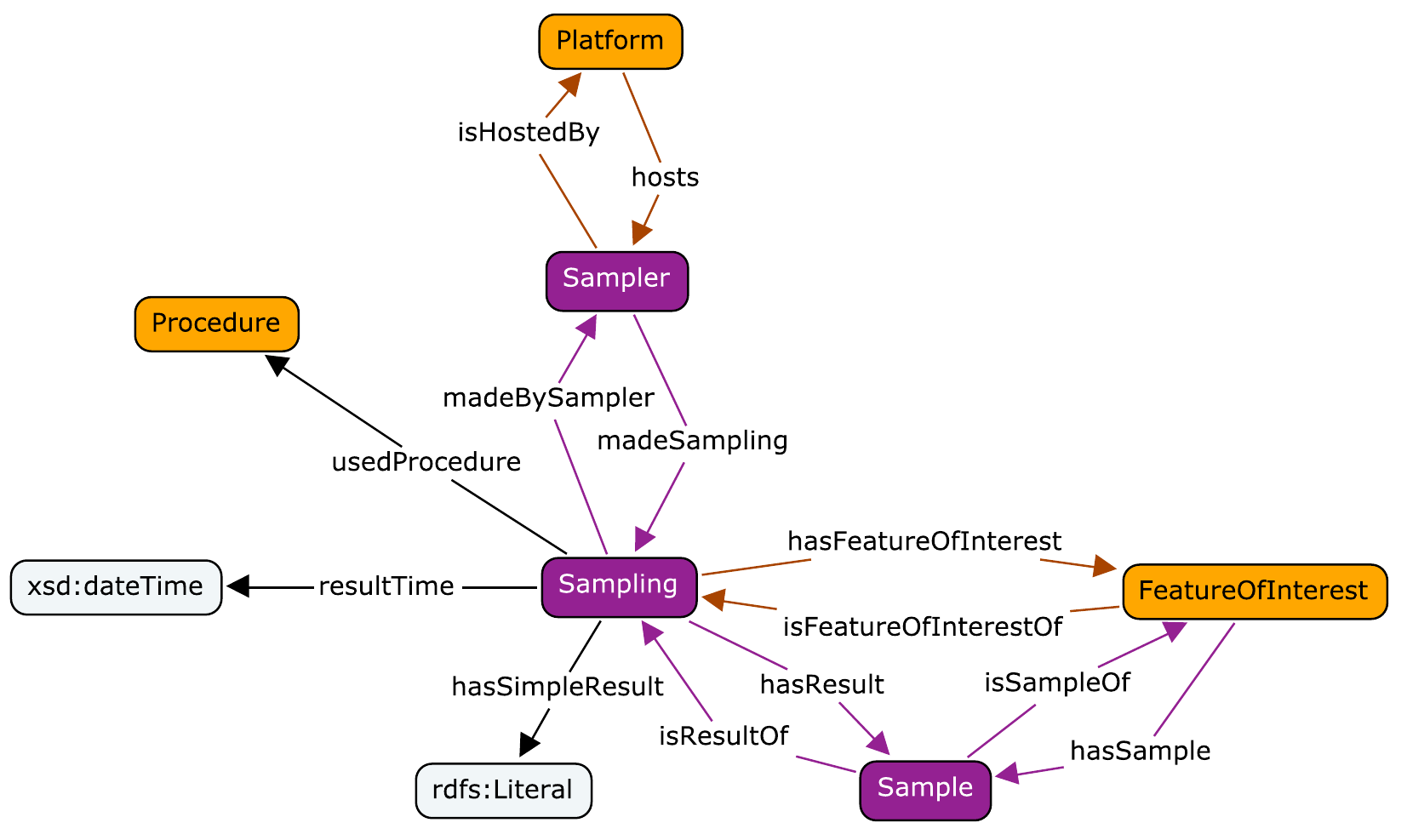}
\caption{Overview of the SOSA Sampling perspective}
\label{fig:sampling}
\end{figure}

A \texttt{Sample} is an object which is representative of a larger object or set of objects, created to support observations. Using a common core structure across different activities was one of the design goals of SOSA; hence, a \texttt{Sampler} is a device, a software or agent that is used by, or implements, a (sampling) procedure to create or transform one or more samples. The act by which a sampler creates a sample is called \texttt{Sampling}, and a sample is also a \texttt{Result}. Samplers can be hosted on platforms, e.g., together with sensors. While each sample is connected to some feature of interest using the \texttt{isSampleOf} relation, a sample may also be a feature of interest itself. 
% SC: Hmm - I think you are reaching a bit far with the following discussion. It is potentially interesting, but I'm not sure the recursivity that you are arguing here applies quite as simply as you say. e.g. a blood sample is representative of the total blood in an organism, but an extract of T-cells taken from that is designed to help characterize only one component of that. I could give similar stories from geology. 
% SC: Suggest omitting the following - For instance, one can create a sample of an ice core sample which was originally extracted using a thermal core drill sampler. In this case the ice core becomes the feature of interest and some fine grained sampler is used to study the core without using it up in the process. We assume that a sample of a sample remains representative of the ultimate feature of interest. Sampling procedures ensure that the sample represents a studied feature and sub-samples of the sample will adhere to the same principle. An exception to this rule would be a case where one wants to study anomalies. 
Fig.~\ref{fig:sampling} depicts the aforementioned classes and their relationships.

Listing~\ref{list_sampling} below shows the result of applying a thermal drill to extract three samples to study the Antarctic ice sheet and then to take $CO_{2}$ observations from one of the samples. It also shows how to locate the sampling event in space and time. While one cannot determine the $CO_{2}$ level for the entire sheet, it is possible to use averages from the sampled values for approximation. Whether this approximation is meaningful for a specific study region and period or the entire sheet depends on our knowledge or theories about the development of Antartica's ice sheet and its uniformity.

\begin{lstlisting}[caption={Sampling modelling},label=list_sampling]
<Antarctic_Ice_Sheet> a sosa:FeatureOfInterest ;
  sosa:hasSample <IceCore12>, <IceCore13>, <IceCore14> .

<IceCore12> a sosa:Sample ;
  sosa:isSampleOf <Antarctic_Ice_Sheet> ;
  sosa:isResultOf <WellDrilling4578> ;
  sosa:madeBySampler <ThermalDrill2> .

<WellDrilling/4578> a sosa:Sampling ;
 geo:lat -73.35 ; 
 geo:long 9.32 ;
 sosa:hasResult <IceCore12> ;
 sosa:madeBySampler <ThermalDrill2> ;
 sosa:resultTime "2017-04-03T11:12:00Z"^^xsd:dateTime ;
 sosa:hasFeatureOfInterest <Antarctic_Ice_Sheet> .

<IceCore12Obs> a sosa:Observation ;
  sosa:hasFeatureOfInterest <IceCore12> ;
  sosa:observedProperty <CO2> ;
  sosa:hasSimpleResult 240 .
\end{lstlisting}

Additional classes and relationships to model relationships between samples are available via the \emph{Sample Relations} vertical segmentation module; see section 5.2 of the SOSA/SSN spec.

\subsection{Actuators and Actuations}
 SOSA also includes classes and relations to model the behaviour of actuation devices, called actuators, that carry out (actuation) procedures to change the state of the world. The modelling of actuations is analogous to the modelling of observations and sampling as it relies on the same core structure.

%\todoah{R3: Similarly, the definition of actuator is that it's a "device" - again, is that restricted to technical devices (e.g. a lightbulb) or can people be actuators? Please clarify.}
%RESPONSE: An actuator can also be a software or agent. We have clarified this in the respective paragraph.

%\todoah{R3: Actuators are stated as responding to an input; however, how this is captured is unclear from Fig 5 and the example listings, as there is no obvious input to the actuator class.  Is the input to the procedure which then triggers the actuation?  The discussion describes how actuators are typically triggered by sensor outputs, given this, it appears that modelling this interaction will be a common task, yet the paper does not provide sufficient clarity about how it should be achieved.}
%RESPONSE: As stated in the paragraph about Actuation the modelling of inputs and outputs for the actuation procedure is outside of the scope of SOSA. An Input and Output class is defined in SSN, but for detailed interface descriptions even SSN relies on external ontologies/data schemes. As mentioned, we have further clarified that in the section where we describe the Procedure class. 

\begin{figure}[ht]
\centering
  \includegraphics[width=\linewidth]{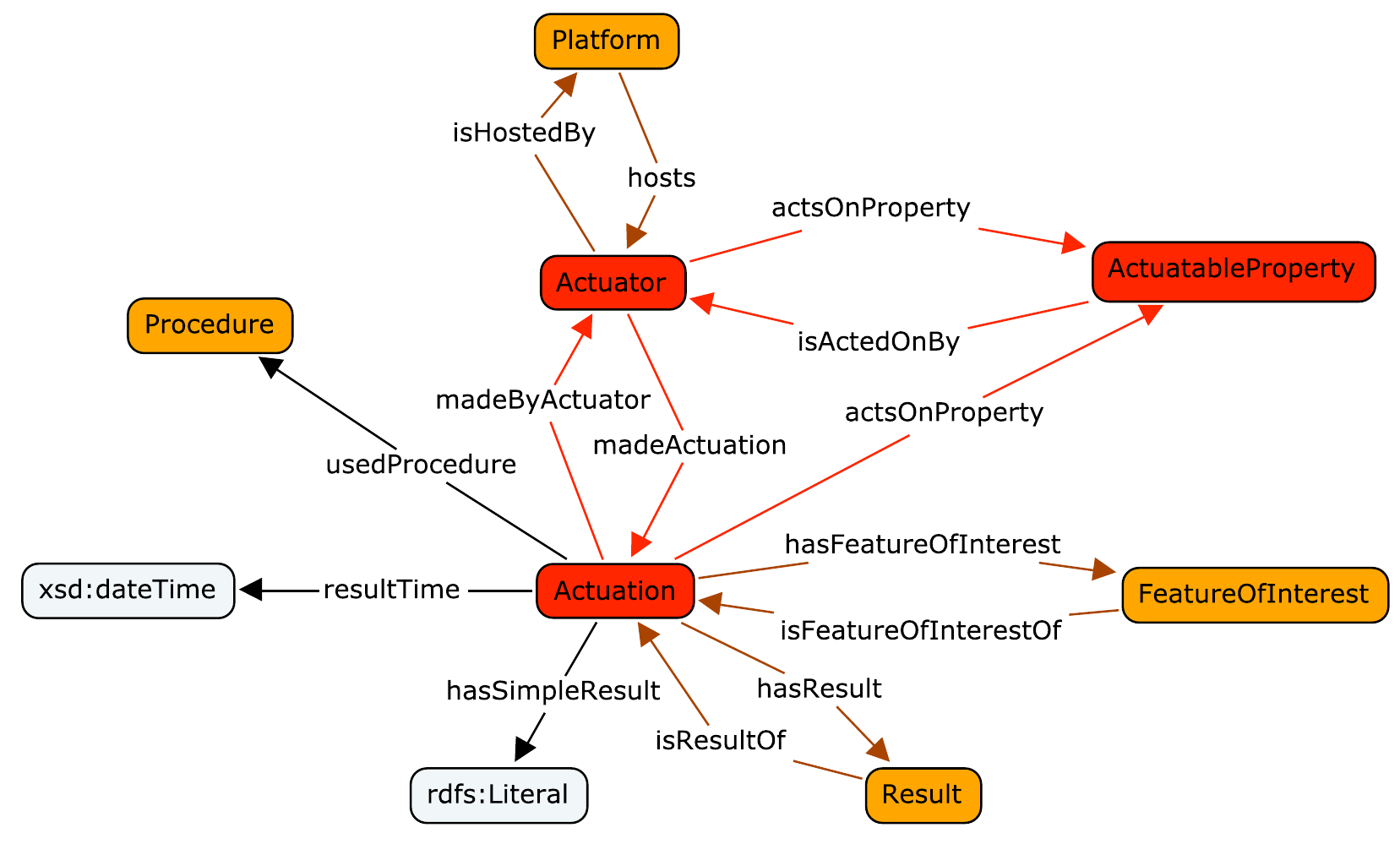}
\caption{Overview of the SOSA Actuation perspective}
\label{fig:actuation}
\end{figure}

 An \texttt{Actuation} is performed by an \texttt{Actuator} and yields a \texttt{Result}. An actuator is a device, software or agent that is used by, or implements, an (actuation) procedure that defines how changes of the state of the world are to be achieved. The actuator responds to an input, defined by the procedure and results in changes in the environment. The set of instructions for turning on and off an Internet of Things enabled light bulb is an example of a procedure. The activity of turning the lightbulb on/off is the actuation and the light bulb (or its socket) is the actuator. The difference with actuations, compared to observations, is that they may be used to model both, a record of how actuations have been performed (a log) and as how to interact with an actuation device (i.e., the procedure how to perform actuations) as well. The former is comparable to the use of the observation class, and the focus of SOSA, while the latter relies on additional axioms provided by the SSN extension to SOSA (i.e., the \texttt{System} concept and its properties \texttt{implements/implementedBy}) as well as on other ontologies and/or specifications that detail the functionality of an actuator further. For example, how much detail is provided to model inputs and outputs of the actuation procedure as well as the orchestration of multiple actuators is beyond the scope of both SOSA and SSN. Existing ontologies such as OWL-S~\cite{owls} and execution frameworks such as WSMX~\cite{Wsmx} can be used together with lower-level specifications such as the W3C Thing Description\footnote{https://w3c.github.io/wot-thing-description/} to model these details. With regards to SOSA, actuators are typically triggered by sensor outputs, i.e., observation results. The thing being acted on is the feature of interest of the actuation and the property being altered is an \texttt{ActuatableProperty}. To close the loop, such actuatable properties are typically also observable properties, e.g., the current state of the aforementioned light bulb. Put differently, the result of an actuation may be the stimulus of a new observation. Fig.~\ref{fig:actuation} depicts the introduced classes and relations.

Listing~\ref{list_actuation} shows how \texttt{Actuation} of an Internet controlled light bulb has been performed. 

\begin{lstlisting}[caption={Actuation modelling},label=list_actuation]
<actuation/046677455286> a sosa:Actuation ;
  sosa:actuatableProperty <philips/046677455286/light> ;
  sosa:hasFeatureOfInterest <light> ;
  sosa:madeByActuator <actuator/philips/HJC42XB/bulb> ;
  sosa:hasResult [
   a qudt-1-1:QuantityValue ;
   qudt-1-1:numericValue "800"^^xsd:double ;
   qudt-1-1:unit qudt-unit-1-1:Lumen ] ;
  sosa:resultTime "2017-10-06T11:26:06Z"^^xsd:dateTime .
\end{lstlisting}

The actuation can be defined to have been made by the actuator (i.e., the light bulb) which also acts as a sensor, as it also allows to read out the current status (ON/OFF) of a light bulb and its current output as measured in lumen. Listing~\ref{list_platform_hosting} shows how the light bulb \lstinline$<actuator/philips/HJC42XB/bulb>$ is hosted by the \emph{Philips Hue Bridge} that acts as a controller for the actuator/sensor, and can be defined as a platform in SOSA.

\begin{lstlisting}[caption={Platform hosting modelling},label=list_platform_hosting]
<philips/46N7743619> a sosa:Platform ;
  rdfs:label "Philips Hue Bridge 46N7743619"@en ;
  rdfs:comment "Philips Hue Bridge - installed in living room"@en ;
  sosa:hosts <actuator/philips/HJC42XB/bulb> ;
  sosa:hosts <sensor/philips/HJC42XB/bulb> .

<actuator/philips/HJC42XB/bulb> a sosa:Actuator ;
  rdfs:label "Philips E27 Bulb - HJC42XB - Turn On/Off"@en ;
  sosa:actsOnProperty <philips/46N7743619/light> ;
  sosa:usedProcedure <philips/46N7743619/switchAPI>.
  
<sensor/philips/HJC42XB/bulb> a sosa:Sensor ;
  rdfs:label "Philips E27 Bulb - HJC42XB - Read Lumen"@en ;
  sosa:observes <philips/46N7743619/light> .
\end{lstlisting}

\section{Vertical Segmentation Alignments}

Vertically segmented modules that import SOSA add higher levels of ontological commitment on top of SOSA by defining new axioms. They are provided for users to either migrate from the old SSN to the new version or to interlink or map data expressed according to standards such as O\&M and OBOE to SSN. SOSA as the core is independent of higher level modules and does not import any other ontologies and its axiomatization is deliberately limited as discussed above. We do not expect users of SOSA to use any of the provided alignment modules (as described in Section 6 of the standard) as they would require the import of an ontology file (which is not common practice in the Schema.org community~\cite{schemaorg}). It is worth noting here, though, that SOSA was developed with these ontologies in mind to act as a common interoperability fallback level between them. An example of that alignment was already given in Listing~\ref{list_jsonld} where we used SOSA in combination with PROV-O.

To give a few more examples of how SOSA relates to other ontologies, \texttt{sosa:Observation} is equivalent to \texttt{o\&m:Observation}, and is a superclass of \texttt{oboe:Measurement}. It should not be confused with \texttt{oboe:Observation}, which is a \emph{collection} of measurements of different properties of the same feature of interest. All of these are, however, subclasses of \texttt{dul:Event} as well as \texttt{prov:Activity}. While a \texttt{sosa:Procedure} can be aligned directly to an \texttt{oboe:Protocol} and as a subclass of \texttt{prov:Plan}, the alignment to an \texttt{o\&m:Process} is defined by an equivalent class relationship to the \emph{Union} of \texttt{sosa:Sensor} and \texttt{sosa-om:ObservationProcedure}, where the latter is a subclass of \texttt{sosa:Procedure} in which every \texttt{usedProcedure$^-$} points to an \texttt{sosa:Observation}. More formally:

{\scriptsize
\begin{align*}
\texttt{ObservationProcedure} \sqsubseteq \;& \texttt{sosa:Procedure}\; \sqcap \\& \forall \texttt{sosa:usedProcedure}^-. \texttt{sosa:Observation}\; \sqcap 	\\& \exists \texttt{sosa:usedProcedure}^-. \texttt{sosa:Observation}\\
\texttt{o\&m:Process}  	\equiv \;& \texttt{sosa:Sensor} \;\sqcup \\& \texttt{ObservationProcedure}
\end{align*}}%

%\todoall{R3: Section 3 discusses the vertical segmentation alignments of SOSA with SSN, DUL, O\&M, OBOE, and SSN-XG ontologies.  I appreciate there are far too many details involved in these alignments to be described in this paper.  While the brief examples provided are helpful, this section does not motivate why these alignments have been defined - what do they allow a user to model that they can't with just using SOSA? Is the purpose of the alignment simply for interoperability with other sensor/observation models? If so, were there any reasons why these specific ones were selected? I'm sure there are good reasons for these alignments, but to users unfamiliar with/new to SOSA, the value of these alignments, why they should be used, and which should be used in different scenarios (e.g. "the alignment with O\&M should be used to do …") are unclear.}
%RESPONSE: We added some sentences in Section 3 that discusses the use of these vertical alignment modules and our expectation that they won't be needed by SOSA users, but that the considered alignment helps community understanding and interlinking once the stricter semantics of SSN are used.

\section{Using SOSA on the Web}
%\todoah{R1: page 6: I am confused by the header of Section 4. How is 4.2 related to Schema.org? Isn't section 4 broader than Schema.org, but about SOSA and other standard vocabularies in general?}
%RESPONSE: We changed the title of the section to ``Using SOSA on the Web'' and included a reference to the Web Data Commons project that provides evidence on the increased use of lightweight vocabularies to express metadata on the Web.

%\todoall{R2: There are lots of different examples used throughout the paper. Obviously, when showing the use of different parts of SOSA examples also need to be different. However, later on, when focusing on the use (e.g. section 4) it is not clear why different examples are used in the listings.}
%RESPONSE: Listing 4.1 is a continuation of our iPhone example in Listing 2.3. Listing 4.2 has a focus on the agent that performed an observation and therefore we have introduced a new example, as the other examples may not necessarily require the modelling of the agent who performed an observation. We have tried to use one example throughout in a previous version of the paper, but it does make the example look fabricated, as few use cases would require observations, actuation and sampling together.

SOSA was engineered to provide a lightweight core vocabulary
to model observations, actuations and samplings and is aimed at a much broader audience than the original SSN, including Web developers who are accustomed to JSON serializations and at most Schema.org style annotations. Considering that already 38\% of all URIs crawled in the Web Data Commons project~\cite{webdatacommons2017} use some form of RDF for metadata modelling, it is expected that practitioners dealing with sensing and actuating devices on the Web will increasingly use metadata to specify the capabilities of these devices and the way that observations and actuations on these devices have been performed.

Schema.org is the de-facto standard vocabulary~\cite{schemaorg} that is embraced by Web developers and helps to integrate data across applications and data formats. Schema.org descriptions can be written using markup attributes in HTML (i.e., using RDFa, Microdata, or JSON-LD as serialization formats). As these serializations provide value for developers and publishers, we will discuss briefly how to use SOSA with Schema.org. 

\subsection{SOSA + Schema.org RDFa}

%\todoah{R3: Section 4 discusses how SOSA metadata can be embedded into HTML as RDFa or encoded as JSON-LD -two formats used for Schema.org descriptors.  Section 2 mentioned the use of Schema.org domainIncludes and rangeIncludes properties, and the RDFa example includes the use of schema:location and associated attributes to describe the location of the observation to define the location that the observation action took place.  This is an interesting example as highlights a correspondence between the SOSA and Schema.org concepts Observation and Action respectively, and it illustrates the additional attributes that annotating a sosa:Observation with type schema:Action introduces - elsewhere in the paper it is implied that observations are associated with a geolocation through that of the platform hosting the sensor that made the observation, while this example illustrates another representation that associated the location directly with an observation.  Are there any preferences or recommendations of best practice for the task of geolocating an observation?}
%RESPONSE: We have now also included a location in the modelling of a Platform in Listing 2.3 and explain in the accompanying paragraph that location can be attached to Observations/Sampling/Actuation as well as to Features of Interest or Sensors/Actuators/Samplers to allow for static as well as remote/mobile application cases.

 RDFa 1.1 provides a set of attribute-level extensions to embed RDF in HTML5 and XHTML5. Revisiting our iPhone example from above, another observation, made a minute later by the same phone, could be modelled using RDFa as described in Listing~\ref{list_rdfa}.

\begin{lstlisting}[caption={RDFa 1.1. serialization},label=list_rdfa]
 <div typeof="sosa:Observation" about="ex:data/observation/346345">
  <div rel="rdf:type" resource="schema:Action"></div>
  <div rel="sosa:hasFeatureOfInterest">
   <div typeof="sosa:FeatureOfInterest" about="ex:earthAtmosphere">
    <div property="rdfs:label" xml:lang="en" content="Earth Atmosphere"></div>
   </div> </div>
  <div rel="sosa:observedProperty" resource="ex:BMP282/AP"></div>
  <div rel="sosa:madeBySensor">
   <div typeof="sosa:Sensor" about="ex:sensor/35-207306-844818-0/BMP282">
    <div rel="sosa:observes" resource="ex:BMP282/AP"></div>
    <div property="rdfs:label" xml:lang="en" content="Bosch Sensortec BMP282"></div>
   </div> </div>
  <div property="sosa:resultTime" datatype="xsd:dateTime" content="2017-06-06T12:37:12+00:00"></div>
  <div property="sosa:hasSimpleResult" datatype="cdt:ucum" content="1022.05 hPa"></div>
  <div property="schema:location" typeof="Place">
   <div property="address" typeof="schema:PostalAddress">
    <div property="schema:addressLocality" xml:lang="en" content="Canberra"></div>
    <div property="schema:addressCountry" xml:lang="en" content="AU"></div>
   </div> </div> </div>
\end{lstlisting}

 The example also integrates the use of Schema.org to further define the location of an observation. Defining a SOSA observation to be also of type Schema.org \texttt{Action} allows the ontology to use the location property for the action to specify the place where the observation took place, e.g., above in Canberra, Australia. A Schema.org action has several other properties that can be used to provide further detail about the observation, including, for example, participants in the action beyond the one who performed it (modelled through the schema:agent relation, which is equivalent to \texttt{sosa:madeBySensor}).

\subsection{SOSA + PROV-O JSON-LD}
\label{sosaprovo}

%\todoah{R3: The JSON-LD example illustrates a JSON-LD encoding of a SOSA + PROV-O example, without use of any Schema.org specific concepts. This example simply presents a serialisation of SOSA/PROV-O that most users will generate automatically through APIs or SPARQL queries, rather than showing any benefit or issues that come from using SOSA with Schema.org.  Overall, rather than providing straightforward serialization of the model, this section would benefit from discussing the pros and cons of using SOSA with Schema.org, any obvious alignments between the two (such as a sosa:Observation also being a schema:Action), and (if the intent of this section is to appeal to web developers currently using Schema.org) why web developers would want to use SOSA concepts rather than sticking with related ones from Schema.org.}
%RESPONSE: The intention of this example was to show the integration of SOSA with PROV-O. Listing 4.1 shows how to integrate SOSA with Schema.org and its mapping to schema:Action (while using a different serialisation, i.e. RDFa). This combination of showing a different seralisiation with two different external vocabularies/ontologies was chosen for space limitations.

 JSON-LD is a JSON-based format to serialize Linked Data. JSON-LD was a reaction to the popularity of JSON, a lightweight, language-independent data interchange format for the Web. It has become the language of choice for the majority of web developers as it is easy to parse and easy to generate. In Listing~\ref{list_jsonld} we show how to serialize a SOSA example in JSON-LD  and how to model an agent and what role he played in the act of sensing, using PROV-O.

%  "@context": {
% "prov": "http://www.w3.org/ns/prov#",
% "sosa": "http://www.w3.org/ns/sosa/",
% "ex": "http://example.org/",
%  },
\begin{lstlisting}[caption={JSON-LD serialization},label=list_jsonld]
 {
  "@graph": [
 {
   "@id": "ex:distancemeter/838725",
   "@type": [ "sosa:Sensor" ],
   "rdfs:label": { "@value": "Leica Disto D2 - 838725"}
 },{
   "@id": "ex:observation/1087",
   "@type": [ "prov:Activity", "sosa:Observation" ],
   "prov:qualifiedAssociation": { "@id": "_:N5bc2a9" },
   "sosa:hasResult": { "@id": "_:Nc90f75" },
   "sosa:madeBySensor": { "@id": "ex:distancemeter/838725"
   },
   "sosa:observedProperty": { "@id": "ex:section/316/length" }
 },{
   "@id": "_:N5bc2a9",
   "@type": "prov:Association",
   "prov:agent": { "@id": "ex:bobthebuilder" },
   "prov:hadRole": { "@id": "ex:structuralEngineer1" },
 }]}
\end{lstlisting}

The example shows an observation of the length of a stretch of road (i.e. \lstinline$ex:section/316/length$) that has been made using the Leica Disto D2 laser distance meter. Since it may be important in this case (for legal/contractual reasons), to record who used the instrument and in which role, the PROV-O ontology can be used to state that ``Bob the Builder'' made \lstinline$ex:observation/1087$ in his role as a structural engineer. Since SOSA has been modelled in an event-centric way, an observation maps to an activity in PROV-O (cf. Section 6.5 in the SSN spec). To associate an agent (\lstinline$ex:bobthebuilder$) and the role (\lstinline$ex:structuralEngineer1$) he played in the activity, a qualified association was used in the example above that uses a blank node (\_:N5bc2a9) for associating the two properties. However, this blank node could also be identified by an IRI instead, e.g. \lstinline$bobAsStructuralEngineer1$, if further statements should be made about that association. Other statements can also be made about the \texttt{prov:Activity}, including when and if another observation was made in the past, who authorized it, and so on.

\section{Modelling with SOSA}
%\todoah{R3: Section 5 discusses three common modelling cases, namely: grouping sets of observations, actuators, or sampling activities together; using individuals or classes; and describing observations as events rather than records.  The discussion in the latter case (section 5.3) provides an important warning for anyone with data represented using the SSN-XG ontology that the different perspectives of SOSA (observation as an event) and SSN-XG (observation as a record) mean that "care must be taking if reasoning over graphs that include data represented using both ontologies," although "reasoning and interpreting" could be a more appropriate phrase in this rather self-evident warning. Is there any advice to users who wish to continue with the observation as an event perspective but also use SOSA - is this possible, what needs to be done?}
%RESPONSE: We have changed the wording as proposed by the reviewer and mentioned that a detailed discussion of how terms in SOSA relate to SSN are outside the scope of this paper and refer the interested reader to the specification.

While we have already provided numerous examples and guidance on modelling with SOSA, this section discusses some common cases that may arise in practice.

\subsection{Collections}
%KJ:I hope you like it :-) AH:Yes, looks good! The only thing we should avoid, or define is "Tbox axioms". The audience may know nothing about TBox or ABox.KJ: ok, I removed the TBOX part
%SC:Yes, that works. Also note that a collection of SOSA observations with a common sosa:hasFeatureOfInterest and sosa:phenomenonTime would be equivalent to an oboe:Observation
There are many reasons to group a set of observations, actuations, or sampling activities together. Avoiding redundancy and reducing storage size is one of them. Given that a sensor may generate thousands of observations during a campaign, generating an assertion for each of these observations to connect them to the used sensor via \texttt{madeBySensor} and to the used procedure via \texttt{usedProcedure} becomes bothersome. While SOSA and SSN do not offer a specific collections class to group observations, actuations, or sampling activities, terminological axioms may take over this role. For example, in case of observations all being taken by a specific sensor, say \texttt{Sensor1}, one would specify a subclass of \texttt{Observation} as follows. 
$$
\texttt{S1Obs} \sqsubseteq \texttt{Observation} \sqcap \exists\texttt{madeBySensor}.
\{\verb|Sensor1|\}.
$$

Instead of connecting every observation with the same sensor, one would simply define observations to be of type S1Obs as shown in Listing~\ref{list_collections}.
\begin{lstlisting}[caption={Collections of Observations},label=list_collections]
<S1Obs1> a sosa:S1Obs ;
  rdfs:label "Observation1 being a Sensor1Observation (S1Obs), i.e., being taken by Sensor1."@en ;
  [...]
<S1Obs2> a sosa:S1Obs ;
  [...]
\end{lstlisting}

 The same applies to procedures, features of interest, observed properties, and so forth, as well as any combination thereof. Classes related to deployments are offered by the SSN module and can be used to group observations by campaigns.
 
 Finally, another question that may arise in the context of storing observations relates to data cataloging and serving. SOSA does not provide classes for information objects, datasets, and catalogs as they are provided by W3C recommendations such as the Data Catalog Vocabulary (DCAT)~\cite{maali2014data}. Similarly, storage, service, and streaming can be handled using methods and tools developed during the initial SSN-XG work  \cite{henson2009semsos,le2011native,janowicz2013restful}.

\subsection{Individuals versus Classes}
Another common issue is the decision between using individuals or classes, e.g., for observed and actuatable properties as well as procedures, to model the interaction with  controlled vocabularies. SOSA does not recommend a specific strategy but relies on either SKOS, language elements such as OWL2's \emph{punning}, or modeling patterns such as casting between individuals (e.g., enumerated literals from code lists) and classes as shown below; see \cite{krisnadhi2015} for details.
\begin{gather*}
  \text{ClassName} \sqsubseteq \exists \text{hasType}.\{\text{classname}\} \\
  \exists \text{hasType}.\{\text{classname}\} \sqsubseteq \text{ClassName} 
\end{gather*}

Returning to the nitrate concentration example given above, a terminological axiom such as
\begin{gather*}
\texttt{cv:NitrateConcentration} \equiv\\ \exists\texttt{sosa:observedProperty}.
\{\texttt{cv:NO3−\_concentration}\}.
\end{gather*}

would be used and also enable the creation of hierarchies, e.g., to say that the \emph{NitrateConcentration} observed property (taken from a code list) is a sub class of \emph{Concentration}.\footnote{See here for an SKOS-based example \url{http://environment.data.gov.au/def/property/nitrate_concentration}.}

%SC: surely, as soon as we have axioms like the guarded domain and range restrictions that SSN adds to SOSA, then individual observations need individual procedures/sensors/fois to serve as the object of corresponding properties? 
%KJ: no :-) We can simply 'cast' them.
%SC: LOL! So much for typed languages (but IANACS so what do I know)

\subsection{Events versus Records}

As alluded to above, perhaps the key conceptual revision compared to the SSO pattern is that observations are now conceived as acts or events. This enabled the establishment of the core structure, with a common pattern for observation, actuation and sampling, which is consistent with other observation models in ontologies, and also makes alignment with PROV-O easy. However, it represents a break with the original SSN ontology that built upon the SSO pattern in which an observation was effectively a record or description of an observation context (a kind of \texttt{dul:Situation}), rather than an activity in the world (a \texttt{dul:Event}) (see  \cite{cox2017omlite} for more discussion). The use of a new namespace for the core classes, as part of SOSA, helps avoid confusion with implementations of the original SSN, but care must be taken if reasoning and interpreting over graphs that includes data represented using both ontologies. In the standardization process we have provided alignment files between the old SSN and SOSA/SSN. A detailed description on how terms in SOSA relate to the old SSN is outside the scope of this paper and we refer the interested reader to the specification~\cite{ssn:17}.

\section{Evidence of use of SOSA}

%\todoall{Section 6: even though the records of usage are available elsewhere, it would be useful to have some references/links here as well, e.g. in the second to last paragraph.}
%RESPONSE: We have included links to the implementations.

The Spatial Data on the Web Working Group collected 51 use cases, from which 62 requirements were derived, which informed the modelling of SOSA and SSN.\footnote{The use cases and requirements can be accessed at \url{https://www.w3.org/TR/sdw-ucr/}.} Based on these requirements, the SOSA ontology was designed to reuse the concepts and properties that have been previously defined in the original SSN ontology, along with additional features specified and requested in previous work such as \cite{OaM} and  \cite{cox2017omlite}. Of the 13 concepts and 21 properties in SOSA, three concepts (i.e. \texttt{Sensor}, \texttt{Platform} and \texttt{FeatureOfInterest}) and eleven properties (i.e. \texttt{hosts}, \texttt{isHostedBy}, \texttt{usedProcedure}, \texttt{hasFeatureOfInterest}, \texttt{madeObservation}, \texttt{madeBySensor}, \texttt{observedProperty}, \texttt{hasResult}, \texttt{isResultOf}, \texttt{phenomenonTime}) are equivalent to a similar term in the original SSN. 
% only 10 properties in this list ... 
New or changed concepts relate to the core structure (\texttt{Procedure}, \texttt{Result}, \texttt{ObservableProperty}, \texttt{ActuatableProperty}), to the expansion of scope to cover sampling and actuation (\texttt{Sampler}, \texttt{Sample}, \texttt{Sampling}, \texttt{Actuator}, \texttt{Actuation}), and to the modified interpretation for \texttt{Observation}.  

The many implementations of the original SSN provide evidence for the utility of those elements carried over from the earlier work, and several implementations using the new elements of SOSA have already been established, including the modelling of samples in Geoscience Australia\footnote{\url{http://pid.geoscience.gov.au/sample/}}, the modelling of oceanographic time series in the South Adriatic Pit (Eastern Mediterranean) by Center for Marine Environmental Sciences, University of Bremen\footnote{\url{https://markusstocker.github.io/eyp-fixo3-ld/browser}}, and a dataset of measurements of a meteorological station by Irstea\footnote{\url{http://ontology.irstea.fr/pmwiki.php/Site/Weather2017}}.

A complete list of the 23 ontologies that already (re)use SOSA and the 23 datasets that use SOSA classes and properties to define data in their applications can be found in the SSN Usage document~\cite{ssn-usage} that records implementation evidence in the SDW as a required part of the W3C standardization process.

\section{Conclusion}
%\todoall{conclusion  1st sentence: 2017and -> 2017 and}
%RESPONSE: Done

With 8.4bn connected things in use worldwide in 2017 % and reaching 20.4bn by 2020\footnote{See \url{https://www.gartner.com/newsroom/id/3598917}} 
and with more than 31\% of websites~\cite{schemaorg} using Schema.org annotations, there is a strong desire by the Web community for a lightweight vocabulary to describe sensors, actuators, and samplers and the acts they can perform. In this article we presented SOSA, a lightweight ontology that represents the core of the new Semantic Sensor Network ontology, a recommendation built in a joint effort by the W3C and OGC, that was specifically designed with the requirements of Web developers, domain scientists, and Linked Data engineers in mind. With its  event-centric view, SOSA aligns well with other standards (i.e., PROV-O, O\&M) and with its Schema.org style \emph{domainIncludes} and \emph{rangeIncludes} annotation properties to provide an informal semantics, the ontology can easily be used in existing applications that support Schema.org annotations to describe IoT devices and their capabilities. The newly established \emph{Spatial Data on the Web Interest Group} will promote the adoption of SOSA and will work on a proper layering (binding) of the newly developed W3C Thing description to SOSA. 

\section*{Acknowledgements}

\small{The SOSA ontology was developed as part of the OGC/W3C Spatial Data on the Web Working Group, see \url{https://www.w3.org/2000/09/dbwg/details?group=75471&public=1} for the list of members. The efforts of W3C staff Phil Archer and Fran\c{c}ois Daoust were invaluable in enabling the successful completion of the work through to publication as a W3C Recommendation and OGC Standard.} The authors acknowledge partial support from NSF (award number 1540849) and  Marie Skodowska-Curie Programme H2020-MSCA-IF-2014 (SMARTER project, under Grant No. 661180).

\section{References}
%\bibliographystyle{plain}
%\bibliography{references}

\end{document}